\documentclass{article}
\usepackage{iclr2015,times}

\usepackage{wrapfig}
\usepackage{natbib}
\usepackage[utf8]{inputenc}
\usepackage{amsmath}
\usepackage{mathtools}
\usepackage{amsfonts}
\usepackage{amsbsy}
\usepackage{amssymb}
\usepackage{graphicx}
\usepackage{subfigure}
\usepackage{psfrag}
\usepackage{epsfig}
\usepackage{multicol}
\usepackage{cite}
\usepackage[center]{caption} 
\usepackage{xcolor} 
\usepackage{textcomp} 
\usepackage{algorithm}
\usepackage{stmaryrd}

\newif\ifarxiv
\arxivtrue
\iclrfinalcopy

\usepackage{bbm}
\usepackage{mathrsfs}
\usepackage{textcomp}
\usepackage{url}

\newcommand {\RR} {\mathbb R}

\def\R{{\mathbb{R}}}

\def\D{{\mathcal{D}}}

\def\X{{\mathcal{X}}}

\def\N{{\mathcal{N}}}

\title{NICE: Non-linear Independent Components Estimation}
\author{Laurent Dinh
\ \  David Krueger
\ \ Yoshua Bengio\thanks{Yoshua Bengio is a CIFAR Senior Fellow.}\\
D\'epartement d'informatique et de recherche op\'erationnelle\\
Universit\'e de Montr\'eal\\
Montr\'eal, QC H3C 3J7 \\
}
\date{}

\begin{document}
\maketitle

\begin{abstract}
We propose a deep learning framework for modeling complex high-dimensional
densities called Non-linear Independent Component Estimation (NICE).  It is based
on the idea that a good representation is one in which the data has a distribution
that is easy to model. For this purpose, a
non-linear deterministic transformation of the data is learned that maps it to a latent space
so as to make the transformed data conform to a factorized distribution,
i.e., resulting in independent latent variables. 
We parametrize this transformation so that computing the determinant of the Jacobian and inverse Jacobian is trivial, yet we maintain the
ability to learn complex non-linear transformations, via a composition of
simple building blocks, each based on a deep neural network. The training
criterion is simply the exact log-likelihood, which is tractable. Unbiased ancestral
sampling is also easy. We show that this approach yields good
generative models on four image datasets and can be used for inpainting.
\end{abstract}

\section{Introduction}

One of the central questions in unsupervised learning is how to capture
complex data distributions that have unknown structure. Deep learning
approaches~\citep{Bengio-2009-book} rely on the learning of a representation
of the data that would capture its most important factors
of variation. This raises the question: {\em what is a good representation?}
Like in recent work~\citep{Kingma+Welling-ICLR2014,Rezende-et-al-arxiv2014,Ozair+Bengio-arxiv2014},
we take the view that {\em a good representation is one in which the distribution
of the data is easy to model}.  In this paper,  we consider the
special case where we ask the learner to find a transformation $h=f(x)$ of the data
into a new space such that the resulting distribution {\em factorizes}, i.e., 
{\em the components $h_d$ are independent}:
\[
  p_H(h) = \prod_d p_{H_d}(h_d).
\]

The proposed training criterion is directly derived from the log-likelihood.
More specifically, we consider a change of variables $h=f(x)$,
which assumes that $f$ is invertible and the dimension of $h$ is the same
as the dimension of $x$, in order to fit a distribution $p_{H}$.
The change of variable rule gives us: 
\begin{equation}
\label{eq:change-var}
  p_X(x) = p_H(f(x)) \lvert \det \frac{\partial f(x)}{\partial x} \rvert.
\end{equation}
where $\frac{\partial f(x)}{\partial x}$ is the Jacobian matrix of function $f$ at $x$.
In this paper, we choose $f$ such that the determinant of the Jacobian is
trivially obtained. Moreover, its inverse $f^{-1}$ is also trivially obtained,
allowing us to sample from $p_X(x)$ easily as follows:
\begin{align}
\label{eq:sampling}
h &\sim~p_H(h) \nonumber \\
x & =~f^{-1}(h)
\end{align}
A key novelty of this paper is the design of such a transformation $f$
that yields these two properties of ``easy determinant of the Jacobian''
and ``easy inverse'', while allowing us to have as much capacity as needed
in order to learn complex transformations. The core idea
behind this is that we can split $x$ into two blocks $(x_1,x_2$) and
apply as building block a transformation from $(x_1,x_2)$ to $(y_1,y_2)$
of the form:
\begin{align}
  y_1=&~x_1 \nonumber \\
  y_2=&~x_2 + m(x_1)
\end{align}
where $m$ is an arbitrarily complex function (a ReLU MLP in our experiments). This building
block has a unit Jacobian determinant for any $m$ and is trivially invertible
since:
\begin{align}
   x_1 =&~ y_1 \nonumber\\
   x_2 =&~ y_2 - m(y_1).
\end{align}
The details, surrounding discussion, and experimental results
are developed below.
\section{Learning bijective transformations of continuous probabilities} 
\begin{figure}
    \centering \subfigure[Inference]{
    \hspace*{16pt}
    \includegraphics[scale=1]{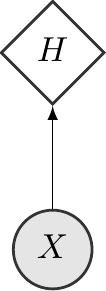}
    \hspace*{16pt}
    \label{fig:inference}}
    \subfigure[Sampling]{
    \hspace*{16pt}
    \includegraphics[scale=1]{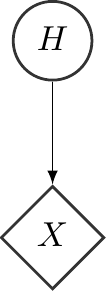}
    \hspace*{16pt}
    \label{fig:sampling}}
    \subfigure[Inpainting]{
    \hspace*{16pt}
    \includegraphics[scale=1]{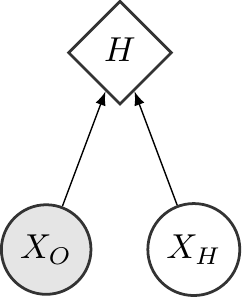}
    \hspace*{16pt}
    \label{fig:inpainting}}
    \caption{Computational graph of the probabilistic model, using the following formulas.\newline
    (a) Inference: $\log(p_X(x)) = \log(p_H(f(x))) + \log(\lvert \det(\frac{\partial f(x)}{\partial x}) \rvert)$\newline
    (b) Sampling: $h \sim p_H(h), x = f^{-1}(h)$\newline
    (c) Inpainting: $\max_{x_H} \log(p_X((x_{O},x_{H}))) = \max_{x_H} \log(p_H(f((x_{O}, x_{H})))) + \log(\lvert \det(\frac{\partial f((x_{O}, x_{H}))}{\partial x}) \rvert)$\newline
    }
\end{figure}


We consider the problem of learning a probability density from a parametric family of
densities $\{p_{\theta}, \theta \in \Theta \}$ over finite dataset $\D$ of $N$
examples, each living in a space $\X$; typically $\X = \RR^{D}$.


Our particular approach consists of learning a continuous, differentiable
almost everywhere non-linear transformation $f$ of the data distribution
into a simpler distribution via {\em maximum likelihood}
using the following change of variables formula:
\begin{align*}
\log(p_X(x)) = \log(p_H(f(x))) +
\log(\lvert \det(\frac{\partial f(x)}{\partial x}) \rvert)
\end{align*}
where $p_H(h)$, the \textit{prior distribution},
will be a predefined density function
\footnote{Note that this prior distribution
does not need to be constant and could also be learned}, 
for example a
standard isotropic Gaussian.
If the prior distribution is factorial (i.e. with independent dimensions), then we obtain the following
\textit{non-linear independent components estimation} (NICE) criterion, which is simply maximum likelihood under our generative model of the data as a deterministic transform of a factorial distribution:
\begin{align*}
\log(p_X(x)) = \sum_{d=1}^{D}{\log(p_{H_{d}}(f_{d}(x)))} + \log(\lvert \det(\frac{\partial f(x)}{\partial x}) \rvert)
\end{align*}
where $f(x) = (f_{d}(x))_{d \leq D}$.

We can view NICE as learning an invertible preprocessing transform of the dataset.
Invertible preprocessings can increase likelihood arbitrarily simply 
by contracting the data.  We use the change of variables formula 
(Eq.~\ref{eq:change-var}) to exactly counteract this phenomenon and use
the factorized structure of the prior $p_H$ to encourage 
the model to discover meaningful structures in the dataset.
In this formula, the determinant of the Jacobian matrix of the transform 
$f$ penalizes contraction and encourages expansion in regions of high density
(i.e., at the data points), as desired.  As discussed in~\citet{Bengio-et-al-ICML2013},
representation learning tends to expand the volume of representation space associated
with more ``interesting'' regions of the input (e.g., high density regions,
in the unsupervised learning case).



In line with previous work with auto-encoders and in particular
the variational auto-encoder~\citep{Kingma+Welling-ICLR2014,
Rezende-et-al-arxiv2014,Mnih+Gregor-ICML2014,Gregor-et-al-ICML2014}, 
we call $f$ the {\em encoder} and its inverse $f^{-1}$
the {\em decoder}. With $f^{-1}$ given, sampling from the model can proceed very easily
by {\em ancestral sampling} in the directed graphical model $H \rightarrow X$, i.e.,
as described in Eq.~\ref{eq:sampling}.

\section{Architecture}
\subsection{Triangular structure}
The architecture of the model is crucial to obtain
a family of bijections whose Jacobian determinant is tractable and whose computation
is straightforward, both forwards (the encoder $f$) and backwards (the decoder $f^{-1}$).
If we use a layered or composed transformation $f = f_L \circ \ldots \circ f_2 \circ f_1$, the
forward and backward computations are the composition of its layers' computations (in the suited order),
and its Jacobian determinant
is the product of
its layers' Jacobian determinants. Therefore we will first aim at
defining those more elementary components.

First we consider affine transformations.
\citep{Rezende-et-al-arxiv2014} and \citep{Kingma+Welling-ICLR2014} provide
formulas for the inverse and determinant when using diagonal matrices, or diagonal
matrices with rank-$1$ correction, as transformation matrices. Another
family of matrices with tractable determinant are triangular
matrices, whose determinants are simply the product of their
diagonal elements. Inverting triangular matrices at test time is 
reasonable in terms of computation. 
Many square matrices $M$ can also be expressed as a product $M = LU$ of upper
and lower triangular matrices. Since such transformations can be composed,
we see that useful components of these compositions include ones whose
Jacobian is diagonal, lower triangular or upper triangular.

One way to use this observation would be to build a neural network with triangular 
weight matrices and bijective activation functions, but this highly constrains
the architecture, limiting design choices to depth and selection of non-linearities.
Alternatively, we can consider a family of functions with triangular
Jacobian. By ensuring that the diagonal elements of the Jacobian are easy to compute,
the determinant of the Jacobian is also made easy to compute.

\subsection{Coupling layer}
\begin{figure}
    \centering \includegraphics[width=.4\textwidth]{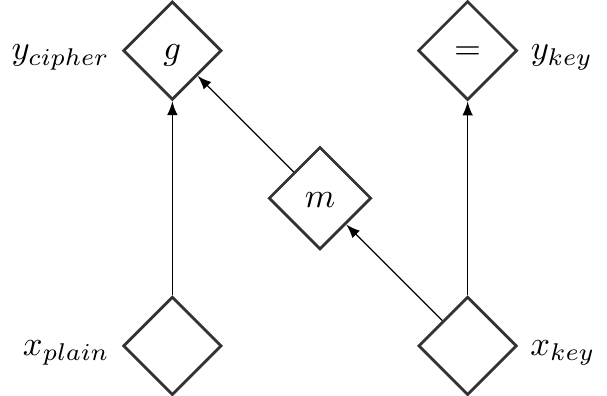}
    \caption{Computational graph of a coupling layer}
    \label{fig:coupling}
\end{figure}

In this subsection we describe a family of bijective transformation with triangular Jacobian therefore tractable Jacobian determinant. That will
serve a building block for the transformation $f$.

\paragraph*{General coupling layer}
Let $x \in \X$, $I_{1}, I_{2}$ a partition of $\llbracket 1, D \rrbracket$
such that $d = \lvert I_{1} \rvert$ and $m$ a function defined on
$\RR^{d}$, we can define $y = (y_{I_{1}}, y_{I_{2}})$ where:
\begin{align*}
y_{I_{1}} &= x_{I_{1}} \\
y_{I_{2}} &= g(x_{I_{2}} ; m(x_{I_{1}}))
\end{align*}
where
$g: \RR^{D-d} \times m(\RR^{d}) \rightarrow \RR^{D-d}$ is the \textit{coupling law},
an invertible map with respect to its first argument given the second. The
corresponding computational graph is shown Fig \ref{fig:coupling}. If we consider
$I_{1} = \llbracket 1, d \rrbracket$ and $I_{2} = \llbracket d, D \rrbracket$,
the Jacobian of this function is:
\begin{align*}
\frac{\partial y}{\partial x} =
\begin{bmatrix}
   I_{d} & 0 \\
   \frac{\partial y_{I_{2}}}{\partial x_{I_{1}}} & \frac{\partial y_{I_{2}}}{\partial x_{I_{2}}} 
\end{bmatrix}
\end{align*}
Where $I_{d}$ is the identity matrix of size $d$. That means that
$\det \frac{\partial y}{\partial x} = \det \frac{\partial y_{I_{2}}}{\partial x_{I_{2}}}$. 
Also, we observe we can invert the mapping using:
\begin{align*}
x_{I_{1}} &= y_{I_{1}} \\
x_{I_{2}} &= g^{-1}(y_{I_{2}} ; m(y_{I_{1}}))
\end{align*}
We call such a transformation a \textit{coupling layer} with
\textit{coupling function} $m$.  

\paragraph*{Additive coupling layer}
For simplicity, we choose as coupling law an
\textit{additive coupling law} $g(a ; b) = a + b$ so that by taking
$a = x_{I_{2}}$ and $b = m(x_{I_{1}})$:
\begin{align*}
y_{I_{2}} &= x_{I_{2}} + m(x_{I_{1}}) \\
x_{I_{2}} &= y_{I_{2}} - m(y_{I_{1}})
\end{align*}
and thus computing the inverse of this
transformation is only as expensive as computing the transformation itself.
We emphasize that there is no restriction placed on the choice of coupling function $m$ (besides having the proper domain and codomain).
For example, $m$ can be a neural network with $d$ input units and $D - d$ output units.


Moreover, since $\det \frac{\partial y_{I_{2}}}{\partial x_{I_{2}}} = 1$, an 
additive coupling layer transformation has a unit Jacobian
determinant in addition to its trivial inverse. One could also
choose other types of coupling, such as a
\textit{multiplicative coupling law} $g(a ; b) = a \odot b,~b \neq 0$ or an
\textit{affine coupling law} $g(a ; b) = a \odot b_{1} + b_{2},~b_{1} \neq 0$ if 
$m: \R^{d} \rightarrow \R^{D-d} \times \R^{D-d}$. We chose the
additive coupling layer for numerical stability reason as the transformation
become piece-wise linear when the {\em coupling function}, $m$, is a rectified
neural network.

\paragraph*{Combining coupling layers}
We can compose several coupling layers to obtain a more complex
layered transformation.
Since a coupling layer leaves part of its input unchanged, we need to
exchange the role of
the two subsets in the partition in alternating layers, so that the composition of two coupling layers 
modifies every dimension. Examining the Jacobian, {\em we observe that at least three coupling
layers are necessary to allow all dimensions to influence one another}. We generally use four.

\subsection{Allowing rescaling}
As each additive coupling layers has unit Jacobian determinant
(i.e. is volume preserving), their
composition will necessarily have unit Jacobian determinant too.
In order to adress this issue, we include
a diagonal scaling matrix $S$ as the top layer, which multiplies the
$i$-th ouput value by $S_{ii}$:
$(x_{i})_{i \leq D} \rightarrow (S_{ii}x_{i})_{i \leq D}$. This allows
the learner to give more weight (i.e. model more variation)
on some dimensions and less in others.

In the limit where $S_{ii}$ goes to $+\infty$ for some $i$,
the effective dimensionality of the data has been reduced by $1$. This is possible so long as
$f$ remains invertible around the data point. With such a scaled diagonal last stage along with lower triangular
or upper triangular stages for the rest (with the identity in their diagonal),
the NICE criterion has the following form:
\begin{align*}
\log(p_X(x)) = \sum_{i=1}^{D}{\left[\log(p_{H_{i}}(f_{i}(x))) + \log(\lvert S_{ii} \rvert)\right]}.
\end{align*}
We can relate these scaling factors to the eigenspectrum of a PCA, showing how much variation
is present in each of the latent dimensions
(the larger $S_{ii}$ is, the less important the dimension $i$ is).  
The important dimensions of the spectrum can be viewed as a manifold learned by the algorithm. 
The prior term encourages $S_{ii}$ to be small, while the determinant
term $\log S_{ii}$ prevents $S_{ii}$ from ever reaching $0$.

\subsection{Prior distribution}
As mentioned previously, we choose the prior distribution to be factorial,
i.e.:
\begin{align*}
p_H(h) = \prod_{d=1}^{D}{p_{H_{d}}(h_{d})}
\end{align*}
We generally pick this distribution in the family of standard distribution,
e.g. gaussian distribution:
\begin{align*}
\log(p_{H_{d}}) = -\frac{1}{2}(h_{d}^{2} + \log(2 \pi))
\end{align*}
or logistic distribution:
\begin{align*}
\log(p_{H_{d}}) = -\log(1 + \exp(h_d)) - \log(1 + \exp(-h_d))
\end{align*}
We tend to use the logistic distribution as it tends to provide a better
behaved gradient.

\section{Related methods}

Significant advances have been made in generative models. Undirected
graphical models like deep Boltzmann machines (DBM) \citep{SalHinton09}
have been very successful and an intense subject of research, due to efficient approximate
inference and learning techniques that these models allowed. However, these
models require Markov chain Monte Carlo (MCMC) sampling procedure for training
and sampling and these chains are generally slowly mixing when the target
distribution has sharp modes. In addition, 
the log-likelihood is intractable, and the best known estimation procedure,
annealed importance sampling (AIS)
\citep{Salakhutdinov+Murray-2008}, might yield an overly optimistic
evaluation \citep{Grosse-et-al-ICML2013}.

Directed graphical models lack the conditional independence structure
that allows DBMs efficient inference.
Recently, however, the development of variational auto-encoders (VAE) 
\citep{Kingma+Welling-ICLR2014,
Rezende-et-al-arxiv2014,Mnih+Gregor-ICML2014,Gregor-et-al-ICML2014} - allowed
effective approximate inference during training.
In constrast with the NICE model, these approaches use
a stochastic encoder $q(h \mid x)$ and an imperfect decoder $p(x \mid h)$,
requiring a reconstruction
term in the cost, ensuring that
the decoder approximately inverts the encoder.
This injects noise into the auto-encoder loop, since $h$ is sampled from $q(h \mid x)$,
which is a variational approximation to the true posterior $p(h \mid x)$.
The resulting training criterion is the variational lower bound on the log-likelihood of the data.
The generally fast ancestral sampling technique that directed graphical models provide make these models appealing. 
Moreover, the importance sampling estimator of the log-likelihood is guaranteed not to be optimistic in expectation. 
But using a lower bound criterion might yield a suboptimal solution with respect to the true log-likelihood.
Such suboptimal solutions might for example inject a significant amount of unstructured noise in the generation
process resulting in unnatural-looking samples.
In practice, we can output a statistic of $p(x \mid h)$, like the expectation
or the median, instead of an actual sample. 
The use of a deterministic decoder can be motivated by the objective of eliminating low-level noise,
which gets automatically added at the last stage of generation in models 
such as the VAE and Boltzmann machines (the visible are considered independent, given the hidden).

The NICE criterion is very similar to the criterion of the variational auto-encoder.
More specifically, as the transformation and its
inverse can be seen as a perfect auto-encoder pair \citep{Bengio-arxiv2014},
the reconstruction term is a constant that can be ignored.
This leaves the Kullback-Leibler divergence term of the variational
criterion: $log(p_H(f(x)))$ can be seen as the prior term, which forces the
code $h=f(x)$ to be likely with respect to the prior distribution, and
$\log(\lvert \det \frac{\partial f(x)}{\partial x} \rvert)$ can be seen as
the entropy term. This entropy term reflects the local volume expansion around
the data (for the encoder), which translates into contraction in the decoder $f^{-1}$. In a
similar fashion, the entropy term in the variational criterion encourages
the approximate posterior distribution to occupy volume, which also translates
into contraction from the decoder.
The consequence of perfect reconstruction is
that we also have to model the noise at the top level, $h$, whereas it is generally handled by the
conditional model $p(x \mid h)$ in these other graphical models.

We also observe that by combining the variational criterion with
the reparametrization trick, \citep{Kingma+Welling-ICLR2014} is
effectively maximizing the joint log-likelihood of the pair $(x, \epsilon)$ in a NICE model
with two affine coupling layers (where $\epsilon$ is the auxiliary noise variable) and gaussian prior, see Appendix \ref{appendix:vae}.


The change of variable formula for probability density functions is
prominently used in inverse transform sampling (which is effectively the
procedure used for sampling here). Independent component analysis (ICA)
\citep{hyvarinen2000independent}, and more specifically its maximum
likelihood formulation, learns an orthogonal transformation of the data,
requiring a costly orthogonalization procedure between parameter updates.
Learning a richer family of transformations was proposed in \citep{Bengio91},
but the proposed class of
transformations, neural networks, lacks in general the structure to make the inference
and optimization practical. \citep{chen2000gaussianization} learns a
layered transformation into a gaussian distribution but in a greedy fashion and it fails to deliver a tractable sampling procedure.

\citep{rippel2013high} reintroduces this idea of learning those transformations
but is forced into a regularized
auto-encoder setting as a proxy of log-likelihood maximization due to the lack of
bijectivity constraint. A more principled
proxy of log-likelihood, the variational lower bound, is used more successfully in
nonlinear independent components analysis \citep{hyvarinen1999nonlinear}
via ensemble learning \citep{roberts2001independent, lappalainen2000nonlinear} and in
\citep{Kingma+Welling-ICLR2014, Rezende-et-al-arxiv2014} using a type of Helmholtz machine \citep{Dayan95}.
Generative adversarial networks (GAN) \citep{Goodfellow-et-al-ARXIV2014} also
train a generative model to transform a simple 
(e.g. factorial) distribution into the data distribution, but do not require an encoder that
goes in the other direction. GAN sidesteps the difficulties of
inference by learning a secondary deep network that discriminates
between GAN samples and data.
This classifier network provides a training signal to the GAN generative model,
telling it how to make its output less distinguishable from the training data.

Like the variational auto-encoders, the NICE model uses an encoder to avoid
the difficulties of inference, but its encoding is deterministic. 
The log-likelihood is tractable and the training procedure does not require any sampling (apart from dequantizing the data). 
The triangular structure used in NICE to obtain tractability is also present in another family of tractable density models, the
neural autoregressive networks~\citep{Bengio+Bengio-NIPS99}, which include as a recent and succesful example the
neural autoregressive density estimator (NADE)~\citep{Larochelle+Murray-2011}.
Indeed, the adjacency matrix in the
NADE directed graphical model is strictly triangular. However the
element-by-element autoregressive schemes make the ancestral sampling procedure computationally
expensive and unparallelizable for generative tasks on high-dimensional data, such as image data.
A NICE model using one coupling layer can be seen as a block version of NADE with two blocks.

\section{Experiments}
\subsection{Log-likelihood and generation}
We train NICE on MNIST \citep{lecun1998mnist}, the Toronto Face Dataset
\footnote{We train on unlabeled data for this dataset.}
(TFD) \citep{Susskind2010}, the Street View House Numbers dataset (SVHN)
\citep{Netzer-wkshp-2011} and CIFAR-10 \citep{Krizhevsky2010tr}. As
prescribed in \citep{Benigno-et-al-NIPS2013-small}, 
we use a dequantized version of the data: 
we add a uniform noise of $\frac{1}{256}$ to the data and
rescale it to be in $[0,1]^{D}$ after dequantization. We add a uniform noise of $\frac{1}{128}$
and rescale the data to be in $[-1,1]^{D}$ for CIFAR-10.

The architecture used is a stack of four coupling layers with a diagonal
positive scaling (parametrized exponentially) $\exp(s)$ for the last stage, and with an approximate
whitening for TFD and exact ZCA on SVHN and CIFAR-10.
We partition the input space between by separating odd ($I_{1}$) and even ($I_{2}$)
components, so the equation is:
\begin{align*}
h^{(1)}_{I_{1}} &= x_{I_{1}}\\
h^{(1)}_{I_{2}} &= x_{I_{2}} + m^{(1)}(x_{I_{1}})
\end{align*}
\begin{align*}
h^{(2)}_{I_{2}} &= h^{(1)}_{I_{2}}\\
h^{(2)}_{I_{1}} &= h^{(1)}_{I_{1}} + m^{(2)}(x_{I_{2}})
\end{align*}
\begin{align*}
h^{(3)}_{I_{1}} &= h^{(2)}_{I_{1}}\\
h^{(3)}_{I_{2}} &= h^{(2)}_{I_{2}} + m^{(3)}(x_{I_{1}})
\end{align*}
\begin{align*}
h^{(4)}_{I_{2}} &= h^{(3)}_{I_{2}}\\
h^{(4)}_{I_{1}} &= h^{(3)}_{I_{1}} + m^{(4)}(x_{I_{2}})
\end{align*}
\begin{align*}
h = \exp(s) \odot h^{(4)}
\end{align*}

The coupling functions $m^{(1)}, m^{(2)}, m^{(3)}$ and $m^{(4)}$ used
for the coupling layers are all deep rectified networks
with linear output units.  We use the same network architecture for each coupling function: five hidden layers of $1000$ units for MNIST,
four of $5000$ for TFD, and four of $2000$ for SVHN and CIFAR-10. 
%

\begin{figure}
    \begin{center}
    \begin{tabular}{| c | c | c | c | c |}
      \hline
      Dataset & MNIST & TFD & SVHN & CIFAR-10 \\ \hline
      \# dimensions & $784$ & $2304$ & $3072$ & $3072$ \\ \hline
      Preprocessing & None & Approx. whitening & ZCA & ZCA \\ \hline
      \# hidden layers & $5$ & $4$ & $4$ & $4$ \\ \hline
      \# hidden units & $1000$ & $5000$ & $2000$ & $2000$ \\ \hline
      Prior & logistic & gaussian & logistic & logistic \\ \hline
      Log-likelihood & $1980.50$ & $5514.71$ & $11496.55$ & $5371.78$ \\
      \hline
    \end{tabular}
    \end{center}
    \caption{Architecture and results. \# hidden units refer to the
    number of units per hidden layer.}
    \label{fig:results}
\end{figure}

\begin{figure}
    \begin{center}
    \begin{tabular}{| c | c | c |}
      \hline
      Model & TFD & CIFAR-10 \\ \hline
      NICE & $5514.71$ & $5371.78$ \\ \hline
      Deep MFA & $5250$ & $3622$ \\ \hline
      GRBM & $2413$ & $2365$ \\
      \hline
    \end{tabular}
    \end{center}
    \caption{Log-likelihood results on TFD and CIFAR-10. Note that the
    Deep MFA number correspond to the best results obtained from
    \citep{tang2012deep} but are actually variational lower bound.}
    \label{fig:mfa-results}
\end{figure}

A standard logistic
distribution is used as prior for MNIST, SVHN and CIFAR-10. A standard normal distribution is used as prior for TFD.

\begin{figure}
    \centering \subfigure[Model trained on MNIST]{
      \includegraphics[width=.4\textwidth]{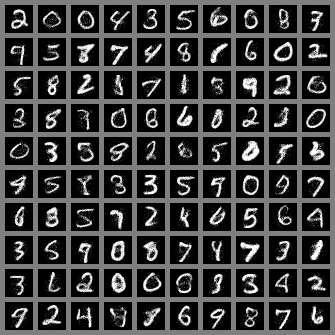} }
    \subfigure[Model trained on TFD]{
      \includegraphics[width=.4\textwidth]{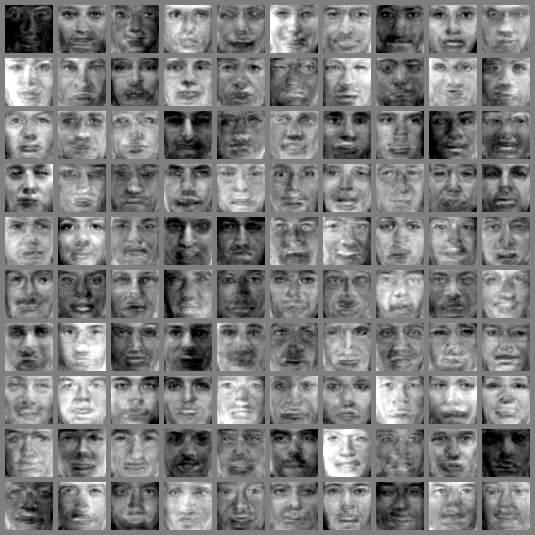} }
    \subfigure[Model trained on SVHN]{
      \includegraphics[width=.4\textwidth]{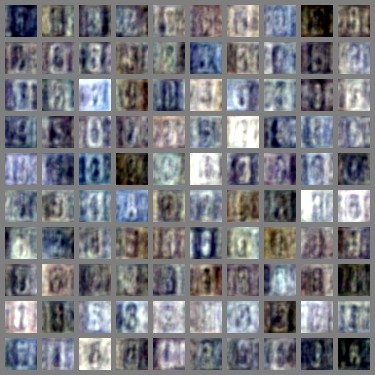} }
    \subfigure[Model trained on CIFAR-10]{
      \includegraphics[width=.4\textwidth]{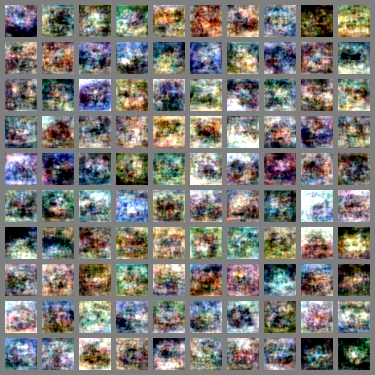} }
    \caption{Unbiased samples from a trained NICE model. We sample
        $h \sim p_{H}(h)$ and we output $x = f^{-1}(h)$.}
    \label{fig:samples}
\end{figure}

The models are trained by maximizing the log-likelihood $\log(p_{H}(h)) + \sum_{i=1}^{D}{s_{i}}$ with AdaM \citep{kingma2014adam} with
learning rate $10^{-3}$, momentum $0.9$, $\beta_{2} = 0.01$,
$\lambda = 1$, and $\epsilon = 10^{-4}$. We select the best model in terms of 
validation log-likelihood after $1500$ epochs.
We obtained a test log-likelihood of $1980.50$ on MNIST, $5514.71$ on TFD,
$11496.55$ for SVHN and $5371.78$ for CIFAR-10. This compares to the best
results that we know of in terms of log-likelihood: $5250$ on TFD and $3622$
on CIFAR-10 with deep mixtures of factor analysers \citep{tang2012deep} (although it is still a lower bound), see Table
\ref{fig:mfa-results}.
As generative models on continuous MNIST are generally evaluated with
Parzen window estimation, no fair comparison can be made.
Samples generated by the trained models are shown in Fig. \ref{fig:samples}.

\subsection{Inpainting}
\begin{figure}
    \centering \subfigure[MNIST test examples]{
      \includegraphics[width=.3\textwidth]{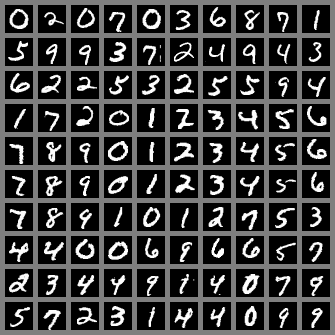} }
    \subfigure[Initial state]{
      \includegraphics[width=.3\textwidth]{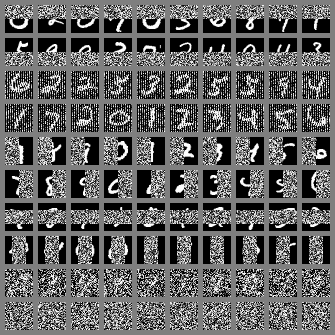} } \subfigure[MAP
      inference of the state]{
      \includegraphics[width=.3\textwidth]{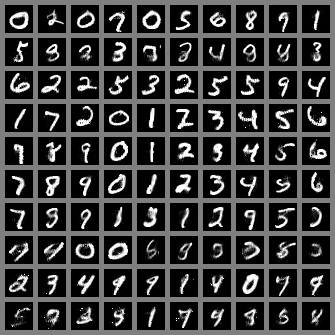} }
    \caption{Inpainting on MNIST. We list below the type of the part of the image
      masked per line of the above middle figure, from top to bottom: top rows, bottom rows, 
      odd pixels, even
      pixels, left side, right side, middle vertically, middle
      horizontally, 75\% random, 90\% random. We clamp the pixels that are
      not masked to their ground truth value and infer the state of the masked
      pixels by projected gradient ascent on the likelihood. Note that with middle masks, there is almost no
      information available about the digit.}
    \label{fig:mnist-inpainting}

\end{figure}

Here we consider a naive iterative procedure to implement inpainting with
the trained generative models. For inpainting we clamp the observed dimensions $(x_O)$ to
their values and maximize log-likelihood with respect to the hidden
dimensions $(X_H)$ using projected gradient ascent (to keep the input in its
original interval of values) with gaussian noise with step size
$\alpha_{i} = \frac{10}{100+i}$, where $i$ is the iteration, following the 
stochastic gradient update:
\begin{align*}
x_{H,i+1} &= x_{H,i} + \alpha_{i}(\frac{\partial \log(p_X((x_{O},x_{H,i})))}{\partial x_{H,i}} + \epsilon)\\
\epsilon &\sim \N(0, I) 
\end{align*}
where $x_{H,i}$ are the values of the hidden dimensions at iteration $i$.
The result is shown on test examples of MNIST, in Fig~\ref{fig:mnist-inpainting}.
Although the model is not trained for this
task, the inpainting procedure seems to yield reasonable qualitative
performance, but note the occasional presence of spurious modes.

\section{Conclusion}
In this work we presented a new flexible architecture for learning a highly non-linear bijective transformation 
that maps the training data to a space where its distribution is factorized, and a framework
to achieve this by directly maximizing log-likelihood. The NICE model features efficient unbiased ancestral
sampling and achieves competitive results in terms of log-likelihood.

Note that the architecture of our model could be trained using other
inductive principles capable of exploiting its advantages, like
toroidal subspace analysis (TSA) \citep{cohen2014learning}.

We also briefly made a connection with variational auto-encoders. 
We also note that NICE can enable more powerful approximate inference allowing
a more complex family of approximate posterior distributions in those models, or a richer family
of priors.

\section*{Acknowledgements} 
We would like to thank Yann Dauphin, Vincent Dumoulin, Aaron Courville, Kyle Kastner, Dustin Webb, Li Yao
and Aaron Van den Oord for discussions and feedback. Vincent Dumoulin provided code for visualization.
We are grateful towards the developers of Theano 
\citep{bergstra+all-Theano-NIPS2011, Bastien-Theano-2012} and Pylearn2 \citep{pylearn2_arxiv_2013},
and for the computational resources provided by Compute Canada and Calcul Québec, and for the
research funding provided by NSERC, CIFAR, and Canada Research Chairs.

\bibliography{strings,ml,aigaion}
\bibliographystyle{apalike}

\newpage
\appendix
\section{Further Visualizations}

\subsection{Manifold visualization}
\begin{figure}
    \centering \subfigure[Model trained on MNIST]{
      \includegraphics[width=.4\textwidth]{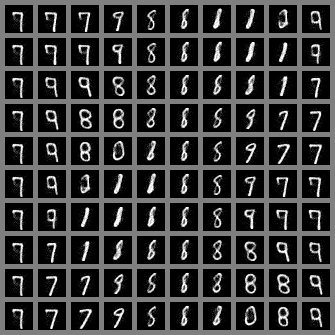} }
    \subfigure[Model trained on TFD]{
      \includegraphics[width=.4\textwidth]{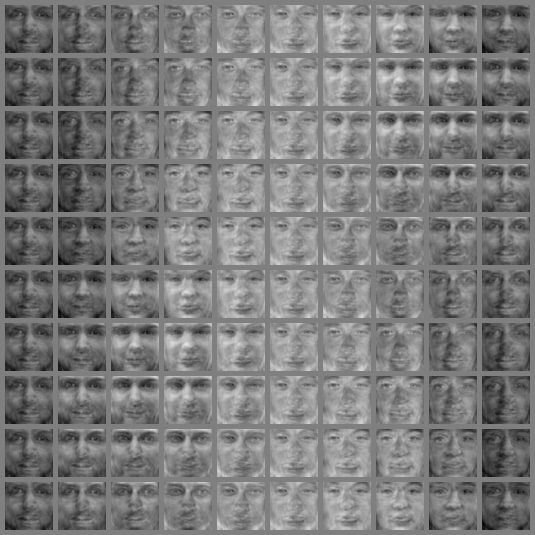} }
    \caption{Sphere in the latent space.
        These figures show part of the manifold structure learned by the model.
        }
    \label{fig:map}
\end{figure}

To illustrate the learned manifold, we
also take a random rotation $R$ of a 3D sphere $S$ in latent space
and transform it to data space, the result $f^{-1}(R(S))$
is shown in Fig \ref{fig:map}.

\subsection{Spectrum}
\begin{figure}[t]
    \centering \subfigure[Model trained on MNIST]{
      \includegraphics[width=.3\textwidth]{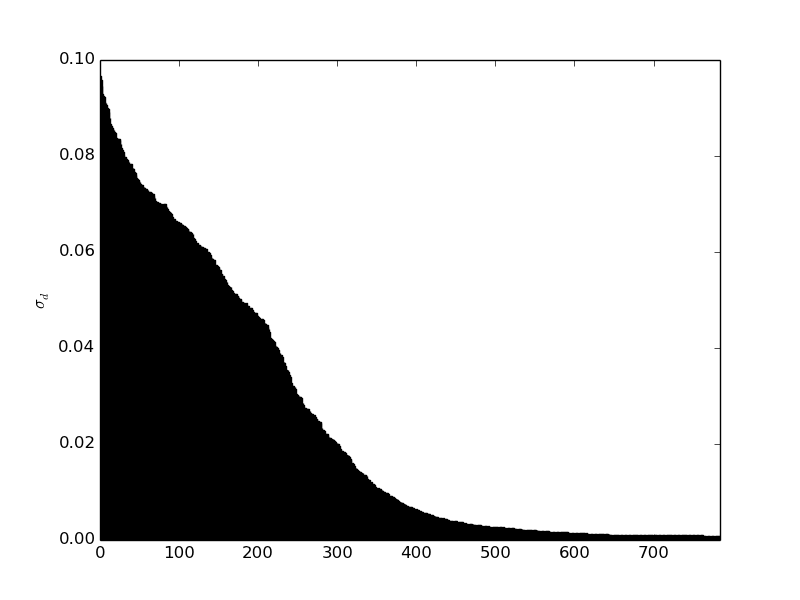} }
    \subfigure[Model trained on TFD]{
      \includegraphics[width=.3\textwidth]{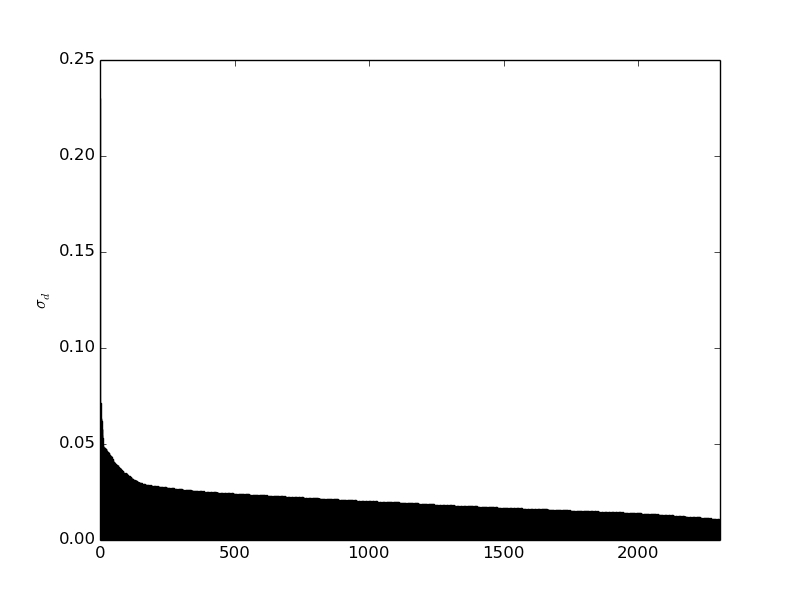} }\\
    \subfigure[Model trained on SVHN]{
      \includegraphics[width=.3\textwidth]{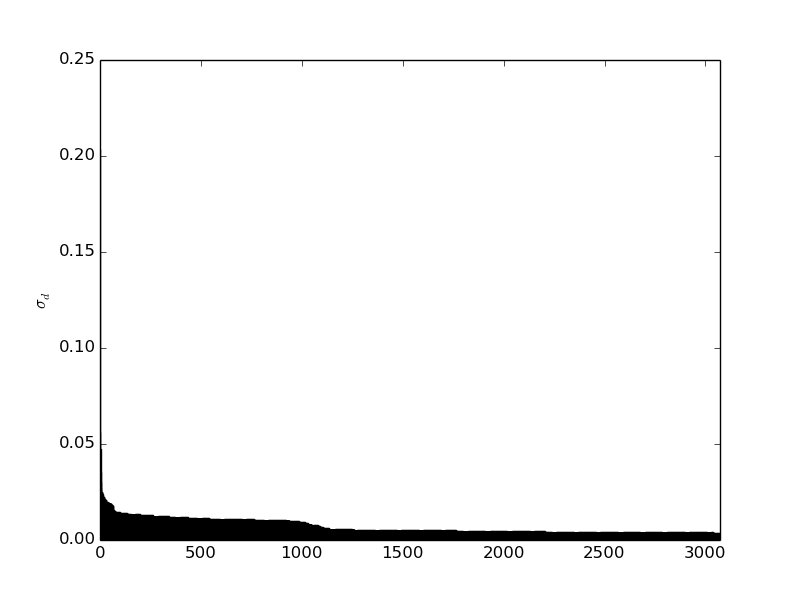} }
    \subfigure[Model trained on CIFAR-10]{
      \includegraphics[width=.3\textwidth]{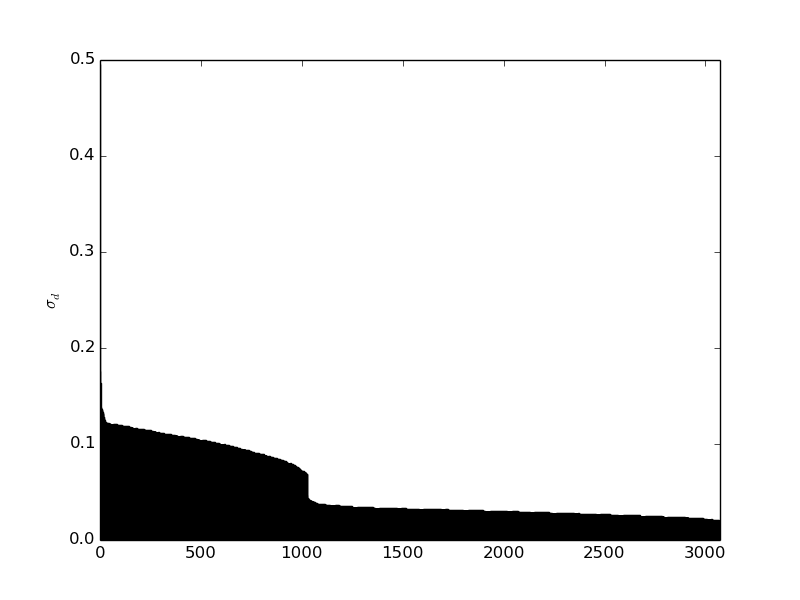} }
    \caption{Decay of $\sigma_{d} = S_{dd}^{-1}$. The large values correspond to dimensions
        on which the model chooses to have larger variations, thus highlighting
        the learned manifold structure from the data. This is the non-linear equivalent
        of the eigenspectrum in the case of PCA. On the x axis are the components $d$ sorted
        by $\sigma_{d}$ (on the y axis).}
    \label{fig:decay}
\end{figure}

We also examined the last diagonal scaling layer and looked at its
coefficients $(S_{dd})_{d \leq D}$. If we consider jointly the prior
distribution and the diagonal scaling layer, $\sigma_{d} = S_{dd}^{-1}$ can
be considered as the scale parameter of each independent component. This shows us
the importance that the model has given to each component
and ultimately how successful the model was at learning manifolds. We sort
$(\sigma_{d})_{d \leq D}$ and plot it in Fig \ref{fig:decay}.

\section{Approximate whitening}
The procedure for learning the approximate whitening is using the NICE framework, with an
affine function and a standard gaussian prior. We have:
\[
z = Lx + b
\]
with $L$ lower triangular and $b$ a bias vector. This is equivalent to learning a gaussian distribution.
The optimization procedure is the same as NICE: RMSProp with early stopping and momentum.

\section{Variational auto-encoder as NICE}
\label{appendix:vae}
We assert here that the stochastic gradient variational Bayes (SGVB) algorithm maximizes the log-likelihood on the pair $(x, \epsilon).
$\citep{Kingma+Welling-ICLR2014} define a recognition network:
\[
z = g_{\phi}(\epsilon \mid x),~\epsilon \sim \N(0, I)
\]

For a standard gaussian prior $p(z)$ and conditional $p(x \mid z)$, we can define:
\begin{align*}
\xi = \frac{x - f_{\theta}(z)}{\sigma}
\end{align*}

If we define a standard gaussian prior on $h = (z, \xi)$.
The resulting cost function is:
\begin{align*}
\log(p_{(x, \epsilon), (\theta, \phi)}(x, \epsilon)) = \log(p_H(h)) - D_{X}\log(\sigma) + \log(\lvert \det \frac{\partial g_{\phi}}{\partial \epsilon}(\epsilon ; x) \rvert)
\end{align*}
with $D_{X} = dim(X)$. This is equivalent to:
\begin{align*}
\log(p_{(x, \epsilon), (\theta, \phi)}(x, \epsilon)) - \log(p_{\epsilon}(\epsilon)) &= \log(p_H(h)) - D_{X}\log(\sigma) + \log(\lvert \det \frac{\partial g_{\phi}}{\partial \epsilon}(\epsilon ; x) \rvert) - \log(p_{\epsilon}(\epsilon))\\
&= \log(p_H(h)) - D_{X}\log(\sigma) - \log(q_{Z \mid X; \phi}(z))\\
&= \log(p_{\xi}(\xi)p_{Z}(z)) - D_{X}\log(\sigma) - \log(q_{Z \mid X; \phi}(z))\\
&= \log(p_{\xi}(\xi)) + \log(p_{Z}(z)) - D_{X}\log(\sigma) - \log(q_{Z \mid X; \phi}(z))\\
&= \log(p_{\xi}(\xi)) - D_{X}\log(\sigma) + \log(p_{Z}(z)) - \log(q_{Z \mid X; \phi}(z))\\
&= \log(p_{X \mid Z}(x \mid z)) + \log(p_{Z}(z)) - \log(q_{Z \mid X; \phi}(z))
\end{align*}
This is the Monte Carlo estimate
of the SGVB cost function
proposed in \citep{Kingma+Welling-ICLR2014}.

\end{document}